# A New Interpolation Approach
# and Corresponding Instance-Based Learning

Shiyou Lian

Xi'an Shiyou University, Xi'an, China

sylian@xsyu.edu.cn

**Abstract**

Starting from finding approximate value of a function, introduces the measure of approximation-degree between two numerical values, proposes the concepts of "strict approximation" and "strict approximation region", then, derives the corresponding one-dimensional interpolation methods and formulas, and then presents a calculation model called "sum-times-difference formula" for high-dimensional interpolation, thus develops a new interpolation approach — ADB interpolation. ADB interpolation is applied to the interpolation of actual functions with satisfactory results. Viewed from principle and effect, the interpolation approach is of novel idea, and has the advantages of simple calculation, stable accuracy, facilitating parallel processing, very suiting for high-dimensional interpolation, and easy to be extended to the interpolation of vector valued functions. Applying the approach to instance-based learning, a new instance-based learning method — learning using ADB interpolation — is obtained. The learning method is of unique technique, which has also the advantages of definite mathematical basis, implicit distance weights, avoiding misclassification, high efficiency, and wide range of applications, as well as being interpretable, etc. In principle, this method is a kind of learning by analogy, which and the deep learning that belongs to inductive learning can complement each other, and for some problems, the two can even have an effect of "different approaches but equal results" in big data and cloud computing environment. Thus, the learning using ADB interpolation can also be regarded as a kind of "wide learning" that is dual to deep learning.

## 1 Introduction

Instance-based learning[1,2] is also called nonparametric approach[3,4]. Instead of establishing a global model of sample data, the approach uses sample data to interpolate directly to achieve objective function approximation. Examples are the experience, and the specific manifestation of a general rule. From the cognitive perspective, the way of learning based on instances is closer to human learning. So, it makes sense to give machines this learning ability. Instance-based learning has been studied for a long time and many achievements have been made (such as *k*-nearest neighbor algorithm, distance weighted nearest neighbor algorithm, locally weighted regression algorithm etc.), but there are still some problems and shortcomings on which (such as high-dimensional interpolation and misclassification). Therefore, instance-based learning as well as corresponding interpolation technique still needs us to continue to research and develop. On the other hand, the current environments of big data and cloud computing undoubtedly provide strong support for instance-based learning and interpolation, and for which also open up a new place to display its prowess. Inspired by the approximate evaluation method of flexible linguistic functions[5, 6] in reference [7], in this paper, we intend to introduce a measure of degree of approximation between numerical values to study the approximate evaluation of numerical functions, and then explores new interpolation approaches based on the degree of approximation and corresponding instance-based learning methods.

## 2 Approximation-Degree, Strict Approximation and Strict Approximation Region

**Definition** 2-1 Let **R** be real number field, $x_0 \in [a, b] \subset \mathbf{R}$, and $[\alpha_0, \beta_0] \subset [a, b]$ be a neighborhood of $x_0$, which is called the approximation region of $x_0$. For $\forall x \in [a, b]$, say $x$ is approximate to $x_0$ if and only if $x \in [\alpha_0, \beta_0]$.

**Definition** 2-2 Let $\mathbf{R}^n$ be $n$-dimensional real vector space, $\boldsymbol{U}=[a_1, b_1]\times[a_2, b_2]\times\ldots\times[a_n, b_n] \subset \mathbf{R}^n$, and $\boldsymbol{x}_0=(x_{1_0}, x_{2_0}, \ldots, x_{n_0}) \in \boldsymbol{U}$. For $\forall \boldsymbol{x}=(x_1, x_2, \ldots, x_n) \in \boldsymbol{U}$, say $\boldsymbol{x}$ is strictly approximate to $\boldsymbol{x}_0$ if and only if components $x_1, x_2, \ldots, x_n$ of $\boldsymbol{x}$ are approximate to components $x_{1_0}, x_{2_0}, \ldots, x_{n_0}$ of $\boldsymbol{x}_0$, respectively, i.e., $x_i \in [\alpha_i, \beta_i] \subset [a_i, b_i]$ ($[\alpha_i, \beta_i]$ is approximation region of $x_{i_0}$), $i$=1,2, … , $n$; and "square" region $[\alpha_1, \beta_1]\times[\alpha_2, \beta_2] \times \ldots \times [\alpha_n, \beta_n] \subset \boldsymbol{U}$ is called the strict approximation region of $\boldsymbol{x}_0$.

In contrast to the strict approximation region in Definition 2-2, we refer to the “circle” region centered on point $\boldsymbol{x}_0$ as the ordinary approximation region of $\boldsymbol{x}_0 \in \boldsymbol{U}$. The relation between the strict approximation region and the ordinary approximation region of the same (2D) point $\boldsymbol{x}_0$ is shown in Figure 2-1. The illustration also shows the relationship between the strict approximation and the ordinary approximation. In fact, the reference [8] has stated: the geometric meaning of "close to point $(x_1, x_2, \ldots, x_n)$ " is different from that of "close to $x_1$ and close to $x_2$ and close to $x_n$".

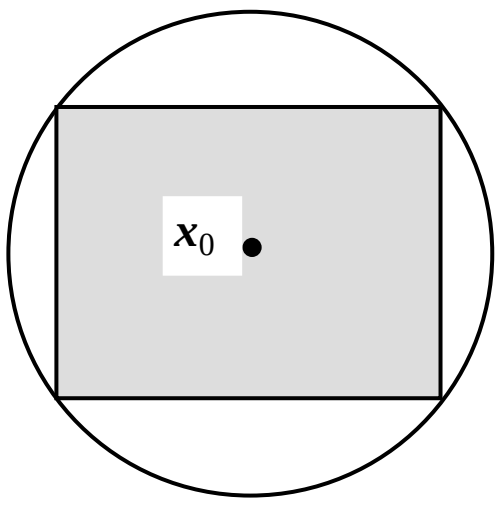

Figure 2-1 An illustration of the relation between strict approximation region and ordinary approximation region

Where the square region is the strict approximation region of point $\boldsymbol{x}_0$, and the circular region is its ordinary approximation region.

**Definition** 2-3 Let **R** be real number field, $x_0 \in [a, b] \subset \mathbf{R}$, and $[\alpha_0, \beta_0] \subset [a, b]$ be the approximation region of $x_0$. Set

$$A_{x_0}(x)=\begin{cases} 1-\frac{x_0-x}{x_0-\alpha_0}=\frac{x-\alpha_0}{x_0-\alpha_0}, & x\in[\alpha_0, x_0] \\ \\ 1-\frac{x-x_0}{\beta_0-x_0}=\frac{x-\beta_0}{x_0-\beta_0}, & x\in[x_0, \beta_0] \end{cases} \qquad (2\text{-}1)$$

to be called the degree of approximation, shortening as approximation-degree, of $x$ to $x_0$. Where $x_0-\alpha_0=r_l$ is called left approximation radius of $x_0$, $\beta_0-x_0=r_r$ is called right approximation radius of $x_0$. We call the function relation defined by the Equation (2-1) to be the approximation-degree function of $x_0$.

## 3 Approximate Evaluation of Functions Based on Approximation-Degree

### 3.1 Finding Approximate Value of a Univariate Function Based on Approximation-Degree

Let **R** be real number field, $U=[a, b]\subset\mathbf{R}$, $V=[c, d]\subset\mathbf{R}$, $y=f(x)$ is a continuous function relation from $U$ to $V$. In the condition that a pair $(x_0, y_0)$ of corresponding values of function $y=f(x)$ and $x'$ approximate to $x_0$ are known, find the approximate value of $f(x')$.

Let the approximation region of $x_0$ be $[a_1, b_1]\subset[a, b]$. According to the definition of the approximation-degree function above, the approximation-degree function of $x_0$ is

$$A_{x_0}(x)=\begin{cases} \frac{x-a_1}{x_0-a_1}, & x\in[a_1, x_0] \\ \\ \frac{x-b_1}{x_0-b_1}, & x\in[x_0, b_1] \end{cases} \qquad (3\text{-}1)$$

And let the approximation region of $y_0$ is $[c_1, d_1]\subset[c, d]$, the approximation-degree function of $y_0$ is

$$A_{y_0}(y)=\begin{cases} \frac{y-c_1}{y_0-c_1}, & y\in[c_1, y_0] \\ \\ \frac{y-d_1}{y_0-d_1}, & y\in[y_0, d_1] \end{cases} \qquad (3\text{-}2)$$

It can be seen that the range of approximation-degree function $A_{x_0}(x)$ is [0, 1] and which is also reversible. In fact, it's easy to obtain that

$$A_{y_0}(y)^{-1}=\begin{cases} d_y(y_0-c_1)+c_1, & d_y\in[0, 1] \\ \\ d_y(y_0-d_1)+d_1, & d_y\in[0, 1] \end{cases} \qquad (3\text{-}3)$$

where $d_y$ is the approximation-degree of $y$ to $y_0$.

Now we find the approximation-degree $A_{x_0}(x')$, and then set $A_{y_0}(y')= A_{x_0}(x')$ (that is, transmitting the approximation-degree of $x'$ (to $x_0$) to $y'$ (to $y_0$)); Further, we derive the required approximate value of $y'$ from approximation-degree $A_{y_0}(y')$ and inverse function $A_{y_0}(y)^{-1}$ of approximation-degree function of $A_{y_0}(y)$.

It can be seen that the inverse function $A_{y_0}(y)^{-1}$ of $A_{y_0}(y)$ is a piecewise function, which has two parallel expressions. Thus, substituting approximation-degree $A_{y_0}(y')=d$ into $A_{y_0}(y)^{-1}$, we can get two $y$ ($y_1$ and $y_2$). Then, which $y$ should be chosen as the desired approximate value of function?

Obviously, the desired $y$ is related to the position of $x'$ relative to $x_0$ and the trend (i.e., being increasing, decreasing, or a constant) of $f(x)$ near $x_0$. Thus, we have the following ideas and techniques:

(1) consider whether the derivative $f'(x_0)$ of the function $f(x)$ at point $x_0$ knows. If the derivative $f'(x_0)$ is known, we can estimate the trend of $f(x)$ near $x_0$ according to the $f'(x_0)$ being positive, negative or zero and then determine the choice of $y$'s value.

(2) consider whether there is a point $x^*$ on the $x'$ side near the point $x_0$ (which does not beyond the approximation region of $x_0$), whose corresponding value of function, $f(x^*) = y^*$, is known. If there is such a point $x^*$, we can estimate the trend of $f(x)$ between the $x_0$ and $x^*$ by utilizing the size relation between the corresponding $y^*$ and $y_0$, and then determine the choice of $y$'s value. For instance, when $x^*<x'<x_0$, if $y^*<y_0$, which then shows that the general trend of function $f(x)$ is increasing on the sub interval $(x^*, x_0)$, thus the $y_1$, i.e., that value less than $y_0$, should be chosen; while if $y^*>y_0$, which then shows that the general trend of function $f(x)$ is decreasing on the sub interval $(x^*, x_0)$, thus the $y_2$, i.e., that value larger than $y_0$, should be chosen.

(3) If the derivative $f'(x_0)$ is unknown and there is no such reference point $x^*$, take the average $\frac{y_1+y_2}{2}$ or take the $y_0$ directly as the approximate value of $f(x')$.

Due to the space limit, in the following, we only discuss the second method further, and use third method to classification problems. As for the first method, it will be introduced in another article.

### 3.2 Finding Approximate Value of a Multivariate Function Based on Approximation-Degree

Let's take the function of two variables as an example to discuss this problem.

Let $z=f(x, y)$ be a function (relation) from $[a_1, b_1]\times[a_2, b_2]$ to $[c, d]$. Suppose a pair of corresponding values, $((x_0, y_0), z_0)$ of function $z=f(x, y)$ and point $(x', y')$ approximate to point $(x_0, y_0)$ are known. In the case that expression of function $f(x, y)$ is unknown or not used, find the approximate value of $f(x', y')$.

By the definition of strict approximation, $(x', y')$ approximate to $(x_0, y_0)$ is equivalent to $x'$ approximate to $x_0$ and $y'$ approximate to $y_0$. Thus, we can find the approximate values $z_x$ and $z_y$ of function $f(x, y_0)$ and $f(x_0, y)$ at points $x'$ and $y'$, respectively. It can be seen that this is really two approximate evaluation problems of univariate functions. Thus, we further imagine that if there is respectively an adjacent point $(x^*, y_0)$ and $(x_0, y^*)$ in the $x$-direction and $y$-direction of the point $(x_0, y_0)$, as shown in Figure 3-1, whose corresponding function values $f(x^*, y_0)$ and $f(x_0, y^*)$ are known, then the approximate value $z_x$ of $f(x', y_0)$ can be got by utilizing $f(x^*, y_0)$, and the approximate value $z_y$ of $f(x_0, y')$ can be got by utilizing $f(x_0, y^*)$, just like that of the previous unary function. Thus, we firstly get separately the approximation-degree $A_{x_0}(x')$ and $A_{y_0}(y')$, then set $A_{z_0}(z) = A_{x_0}(x')$, and $A_{z_0}(z) = A_{y_0}(y')$; and then, substitute them separately into inverse function $A_{z_0}(z)^{-1}$ of $A_{z_0}(z)$ and get two pairs of candidate approximations, then taking separately $f(x^*, y_0)$ and $f(x_0, y^*)$ as reference, choose $z_x$ and $z_y$ from respective candidate values (as shown in Figure 3-1).

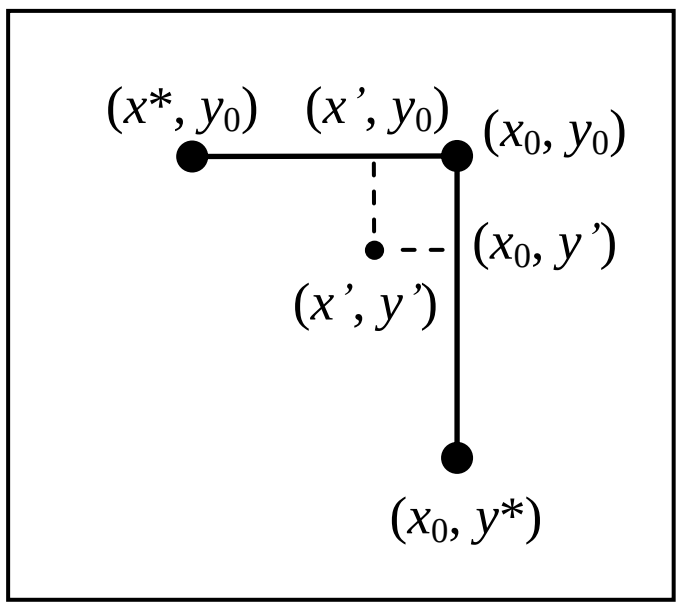

Figure 3-1 Utilizing the values of the function at points ($x^*$, $y_0$) and ($x_0$, $y^*$) to determine the approximate values of $f(x', y_0)$ and $f(x_0, y')$, respectively

Having got the approximate values $z_x$ and $z_y$, how can we further get the approximate value $z$ we need?

Let $z_1=(z_x+z_y)/2$ be the average of $z_x$ and $z_y$. It can be seen from Figure 3-2 that $z_1$ can actually be viewed as an approximate value of $f(x, y)$ at midpoint (denoted by ($x_1$, $y_1$)) between ($x'$, $y_0$) and ($x_0$, $y'$). We can see from the figure that $z_1<z_0$, i.e., the varying trend of function values from $z_0$ to $z_1$ is decreasing. Set $z_0–z_1=c_1$ ($c_1$ is the length of segment $BC$ in Figure 3-3(a)), then $z_1=z_0–c_1$. Then, according to the varying trend of function values from $z_0$ to $z_1$ (i.e., the slope of segment $AB$ in Figure 3-3(a)), also, taking into account that point ($x_1$, $y_1$) is just the midpoint of segment joining points ($x'$, $y'$) and ($x_0$, $y_0$), that is, $\|(x',y')-(x_0,y_0)\|=2\bullet\|(x_1,y_1)-(x_0,y_0)\|$, so we infer that the approximate value of function at point ($x'$, $y'$) can be $z_0–2c_1$ (as shown in Figure 3-3(a)). Thus, it follows that

$$z=z_0-2c_1=z_0-2(z_0-z_1)=z_0-2[z_0-(z_x+z_y)/2] = z_x+z_y-z_0$$

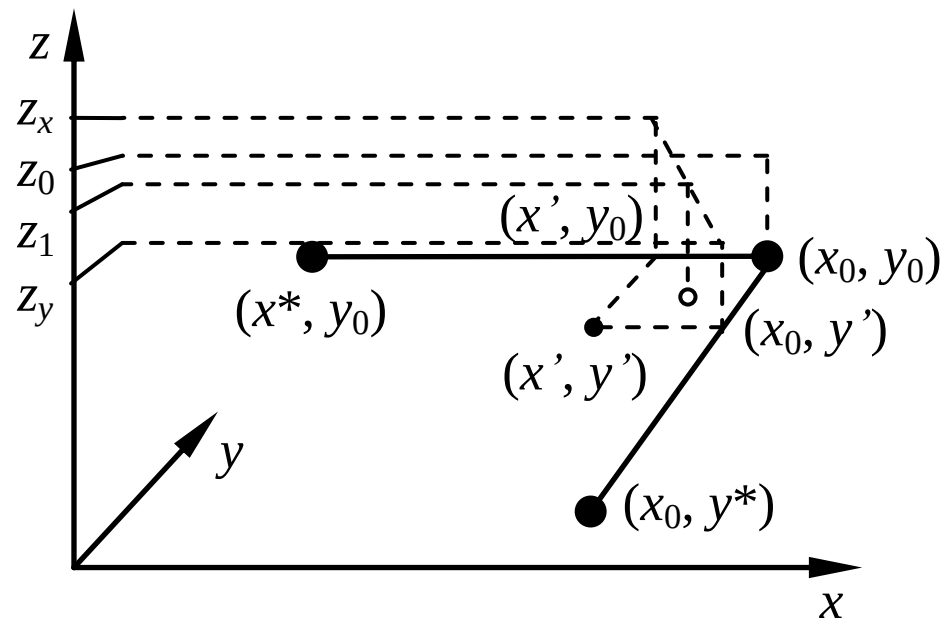

Figure 3-2 Illustration-1 of synthesizing $z_x$ and $z_y$ into $z$

Of course, $z_1$ may also be greater than $z_0$ or equal to $z_0$. If $z_1>z_0$, then the varying trend of function values from $z_0$ to $z_1$ is increasing (as shown in Figure 3-3(b)). Set $z_0–z_1=c_2$ ($c_2$ is the length of segment $BC$ in Figure 3-3(b)), then $z_1=z_0–c_2$. Then, according to the varying trend of function values from $z_0$ to $z_1$ (i.e., the slope of segment $AB$ in Figure 3-3(b)), we infer that the value of function at point ($x'$, $y'$) can be $z_0–2c_2$. Thus,

$$z=z_0+2c_2=z_0+2(z_1-z_0)=z_0+2[\ (z_x+z_y)/2-z_0] = z_x+z_y-z_0$$

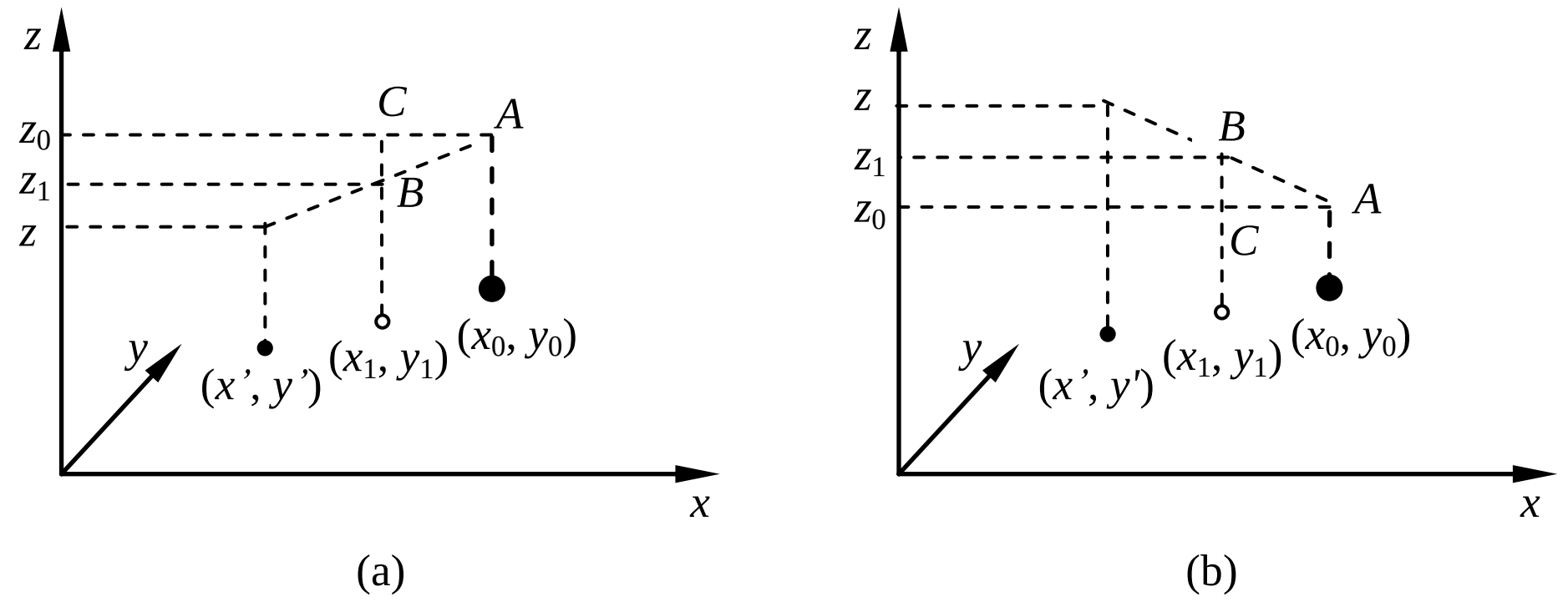

(a) (b)

Figure 3-3 Illustration-2 of synthesizing $z_x$ and $z_y$ into $z$

The third case: $z_1$=$z_0$. This indicates that the values of function remain unchanged from $z_0$ to $z_1$. Thus, we can take $z$=$z_0$. And by $z_0$=$z_1$= ($z_x$+$z_y$)/2, it follows that $2z_0$=$z_x$+$z_y$. Thus,

$$z=2z_0-z_0=z_x+z_y-z_0$$

In summary, we see that, no matter what relationship may be between the average of $z_x$ and $z_y$ and the $z_0$, or no matter how the value of the function varies from $z_0$ to $z_1$, the approximate value of the function at point ($x'$, $y'$) can always be taken as

$$z=z_x+z_y-z_0 \quad (3\text{-}4)$$

This equation is also the calculation model of the approximate value of function of two variables, $z=f(x, y)$.

It can be seen that this is actually decomposing the approximate evaluation of a function of two variables into the approximate evaluation of two functions of one variable, firstly, then, synthesizing two obtained approximate values into a value as an approximate value of the original function of two variables. Extending this technique of "first splitting then synthesizing" to the approximate evaluation of a function of 3 variables, $u= f(x, y, z)$, we obtain the formula synthesizing approximate value of the function is

$$u = u_x+u_y+u_z-2u_0 \quad (3\text{-}5)$$

And then, for a function of $n$ variables, $y= f(x_1, x_2, \dots, x_n)$, the corresponding formula synthesizing approximate value of the function is

$$y= y_{x_1}+y_{x_2}+\cdots+y_{x_n}-(n-1)y_0$$

or

$$y=\sum_{i=1}^{n} y_{x_i}-(n-1)y_0 \quad (3\text{-}6)$$

For convenience of narration, we may as well refer to the Equations (3-4), (3-5) and (3-6) as the **sum-times-difference formula**.

## 4 Interpolation Based on Approximation-Degree

Let $y=f(x)$ be a function (relation) from [$a$, $b$] to [$c$, $d$]. A set of pairs of corresponding values of function $y=f(x)$, {($x_1$, $y_1$), ($x_2$, $y_2$), … , ($x_n$, $y_n$)}, is known, where $x_1<x_2<, \dots, <x_n$. Now the question is: in the case that the expression of function $f(x)$ is unknown or not used, construct an interpolating function $g(x)$ such that $g(x_i)= f(x_i)$ (i=1, 2, … , n), and for other $x\in[a, b]$, $g(x)\approx f(x)$. This is the usual interpolation problem. We now use the approach that finding approximate value of a function above to solve the interpolation problem.

Let $a=x_1$, $x_n=b$, then $x_1, x_2, \dots, x_n$ is a group of interpolation base points (or nodes). We definite the approximation region of $x_1$ as $[x_1, x_2]$, the approximation region of $x_i$ as $[x_{i-1}, x_{i+1}]$ ($i$=2, 3, … , $n$–1), and the approximation region of $x_n$ as $[x_{n-1}, x_n]$, and then definite separately the approximation-degree functions of base point $x_1$, $x_i$, and $x_n$ as

$$A_{x_1}(x)=\frac{x-x_2}{x_1-x_2},\quad x\in[x_1,\ x_2] \tag{4-1}$$

$$A_{x_i}(x)=\begin{cases}\dfrac{x-x_{i-1}}{x_i-x_{i-1}}, & x\in[x_{i-1}, x_i]\\[2ex] \dfrac{x-x_{i+1}}{x_i-x_{i+1}}, & x\in[x_i,\ x_{i+1}]\end{cases} \tag{4-2}$$

$$A_{x_n}(x)=\frac{x-x_{n-1}}{x_n-x_{n-1}},\quad x\in[x_{n-1}, x_n] \tag{4-3}$$

Note that it is not hard to see from the above expressions of approximation-degree function that when $x\in[x_1,\ \frac{x_1+x_2}{2}]$, $[\frac{x_{i-1}+x_i}{2},\ \frac{x_i+x_{i+1}}{2}]$, or $[\frac{x_{n-1}+x_n}{2}, x_n]$, the corresponding approximation-degrees $A_{x_1}(x)$, $A_{x_i}(x)$, and $A_{x_n}(x)$ are always ≥0.5. This means that $x$ is closer to the corresponding base point $x_1$, $x_i$ or $x_n$.

We then define the approximation-degree functions of $y_i$ ($i$=1, 2, … , $n$) in the same principle and way.

$$A_{y_i}(y)=\frac{y-y_{i-1}}{y_i-y_{i-1}}=\frac{1}{y_i-y_{i-1}}y-\frac{y_{i-1}}{y_i-y_{i-1}},\quad y\in[y_{i-1},\ y_i] \tag{4-4}$$

$$A_{y_i}(y)=\frac{y_{i-1}-y}{y_{i-1}-y_i}=\frac{1}{y_i-y_{i-1}}y-\frac{y_{i-1}}{y_i-y_{i-1}},\quad y\in[y_i,\ y_{i-1}] \tag{4-5}$$

$$A_{y_i}(y)=\frac{y_{i+1}-y}{y_{i+1}-y_i}=\frac{1}{y_i-y_{i+1}}y-\frac{y_{i+1}}{y_i-y_{i+1}},\quad y\in[y_i,\ y_{i+1}] \tag{4-6}$$

$$A_{y_i}(y)=\frac{y-y_{i+1}}{y_i-y_{i+1}}=\frac{1}{y_i-y_{i+1}}y-\frac{y_{i+1}}{y_i-y_{i+1}},\quad y\in[y_{i+1},\ y_i] \tag{4-7}$$

Obviously, in the four expressions above, (4-4) = (4-5) and (4-6) = (4-7). Thus, the 4 functional expressions can be reduced as two expressions:

$$A_{y_i}(y)\ =\ \frac{1}{y_i-y_{i-1}}y-\frac{y_{i-1}}{y_i-y_{i-1}}$$

$$A_{y_i}(y)=\frac{1}{y_i-y_{i+1}}y-\frac{y_{i+1}}{y_i-y_{i+1}}$$

And then, we get the inverse expressions of these two functional expressions:

$$A_{y_i}(y)^{-1}=d_y(y_i-y_{i-1})+y_{i-1} \tag{4-8}$$

$$A_{y_i}(y)^{-1}=d_y(y_i-y_{i+1})+y_{i+1} \tag{4-9}$$

Here $d_y=A_{y_i}(y)\in[0, 1]$ is the approximation-degree of $y$ to $y_i$.

Now, set

$$d_y=d_x=A_{x_i}(x)$$

Also, considering that on the sub interval $[\frac{x_{i-1}+x_i}{2},\ x_i]$, the interpolated function $y=f(x)$ may being increasing, decreasing, or a constant; while when $y=f(x)$ is increasing, certainly $y_{i-1}<y_i$, so the corresponding $y\in[\frac{y_{i-1}+y_i}{2},\ y_i]$; when $y=f(x)$ is decreasing, certainly $y_i<y_{i-1}$, so the corresponding $y\in[y_i,\ \frac{y_{i-1}+y_i}{2}]$; and when $y=f(x)$ is a constant, $y_i=y_{i-1}$, so the corresponding $y=y_i\in[\frac{y_{i-1}+y_i}{2},\ y_i]$ as well as $y\in[y_i,\ \frac{y_{i-1}+y_i}{2}]$. Thus, when $x\in[\frac{x_{i-1}+x_i}{2},\ x_i]$, the corresponding $y_i$ and $y_{i-1}$ are adjacent, and here $A_{x_i}(x)=\frac{x-x_{i-1}}{x_i-x_{i-1}}$, thus the Expression (4-8), the expression of the corresponding inverse function $A_{y_i}(y)^{-1}$, becomes

$$A_{y_i}(y)^{-1}=\frac{x-x_{i-1}}{x_i-x_{i-1}}(y_i-y_{i-1})+y_{i-1}=\frac{y_i-y_{i-1}}{x_i-x_{i-1}}x+\frac{x_iy_{i-1}-x_{i-1}y_i}{x_i-x_{i-1}}$$

namely

$$y=\frac{y_i-y_{i-1}}{x_i-x_{i-1}}x+\frac{x_iy_{i-1}-x_{i-1}y_i}{x_i-x_{i-1}},\quad x\in[\frac{x_{i-1}+x_i}{2},\ x_i] \tag{4-10}$$

Similarly, the Expression (4-9) of the inverse function $A_{y_i}(y)^{-1}$ becomes

$$y=\frac{y_i-y_{i+1}}{x_i-x_{i+1}}x+\frac{x_iy_{i+1}-x_{i+1}y_i}{x_i-x_{i+1}},\quad x\in[x_i,\ \frac{x_i+x_{i+1}}{2}] \tag{4-11}$$

Thus, with the two expressions, we can obtain directly the corresponding $y\in[\frac{y_{i-1}+y_i}{2},\ y_i]$ or $[y_i,\ \frac{y_{i-1}+y_i}{2}]$ from $x\in[\frac{x_{i-1}+x_i}{2},\ x_i]$, and obtain directly the corresponding $y\in[y_i,\ \frac{y_i+y_{i+1}}{2}]$ or $[\frac{y_i+y_{i+1}}{2},y_i]$ from $x\in[x_i,\ \frac{x_i+x_{i+1}}{2}]$.

Actually，Equations (4-10) and (4-11) are two interpolation formulas. In this way, we actually derive an interpolation approach by using the approximate evaluation of function based on approximation-degree. We call this approach to be the approximation-degree-based interpolation, or **ADB interpolation** for short.

Specifically, the practice of ADB interpolation is: take base points $a=x_1,x_2,\dots,x_n=b$ as points of view, according to base points and their approximation regions to partition interval $[a,\ b]=[x_1,\ x_n]$ into $2n-2$ subintervals as shown in Figure 4-1, $[x_1,\frac{x_1+x_2}{2}]$, $[\frac{x_1+x_2}{2},x_2]$, $[x_2,\frac{x_2+x_3}{2}]$,…,$[\frac{x_{n-1}+x_n}{2},\ x_n]$, as interpolation intervals; then, for evaluated point $x\in[\frac{x_{i-1}+x_i}{2},\ x_i]$ do interpolating with Formula (4-10), for evaluated point $x\in[x_i,\ \frac{x_i+x_{i+1}}{2}]$ do interpolating with Formula (4-11). Since each specific interpolating formula implies the trend of interpolated function $y=f(x)$ on the corresponding subintervals, therefore, there is no much error between the obtained approximate value and the expected value, and there would not occur the case that two $y$-values are got from an $x$.

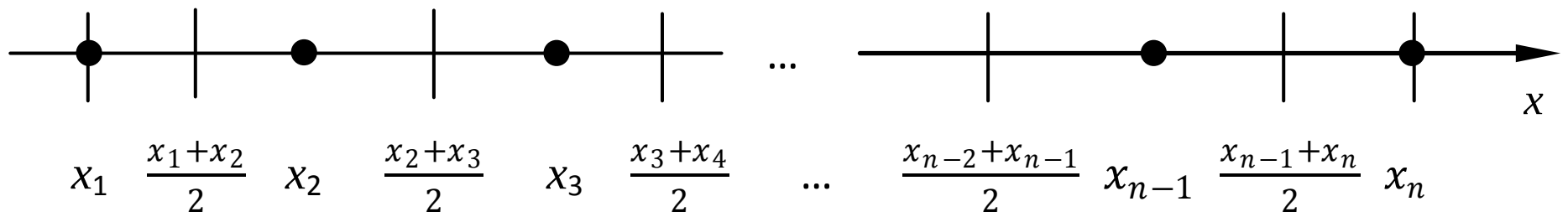

Figure 4-1 Illustration of interpolation intervals partitioned by base points and their approximation domains in one-dimensional interpolation

**Example** 4-1 Use ADB interpolation to do interpolation for functions $y$=sin $x$.

We take sampled data points of $y$=sin $x$, $(x_i, y_i)$, as follows:

$x_i$: 0×π , 0.1×π , 0.2×π , 0.3×π , … , 1.9×π , 2×π 。

$y_i$: sin($x_i$)。

and take interpolation points,

$x$: 0×π, 0.02×π, 0.04×π, 0.06×π, … , 1.98×π, 2×π.

Using our ADB interpolation to do interpolation, the corresponding $y$-values obtained are as follows:

| | | | | | | | | | |
|---|---|---|---|---|---|---|---|---|---|
| 0 | 0.0624 | 0.1249 | 0.1873 | 0.2497 | 0.3118 | 0.3681 | 0.4245 | 0.4808 | 0.5371 |
| 0.5923 | 0.6369 | 0.6816 | 0.7263 | 0.7710 | 0.8133 | 0.8420 | 0.8707 | 0.8994 | 0.9281 |
| 0.9530 | 0.9629 | 0.9728 | 0.9827 | 0.9926 | 0.9975 | 0.9876 | 0.9778 | 0.9679 | 0.9580 |
| 0.9424 | 0.9138 | 0.8851 | 0.8564 | 0.8277 | 0.7934 | 0.7487 | 0.7040 | 0.6593 | 0.6146 |
| 0.5653 | 0.5089 | 0.4526 | 0.3963 | 0.3400 | 0.2809 | 0.2185 | 0.1561 | 0.0936 | 0.0312 |
| −0.0312 | −0.0936 | −0.1561 | −0.2185 | −0.2809 | −0.3400 | −0.3963 | −0.4526 | −0.5089 | |
| −0.5653 | −0.6146 | −0.6593 | −0.7040 | −0.7487 | −0.7934 | −0.8277 | −0.8564 | −0.8851 | −0.9138 |
| −0.9424 | −0.9580 | −0.9679 | −0.9778 | −0.9876 | −0.9975 | −0.9926 | −0.9827 | −0.9728 | −0.9629 |
| −0.9530 | −0.9281 | −0.8994 | −0.8707 | −0.8420 | −0.8133 | −0.7710 | −0.7263 | −0.6816 | −0.6369 |
| −0.5923 | −0.5371 | −0.4808 | −0.4245 | −0.3681 | −0.3118 | −0.2497 | −0.1873 | −0.1249 | −0.0624 |
| 0.0000 | | | | | | | | | |

And the effect is shown in Figure 4-2.

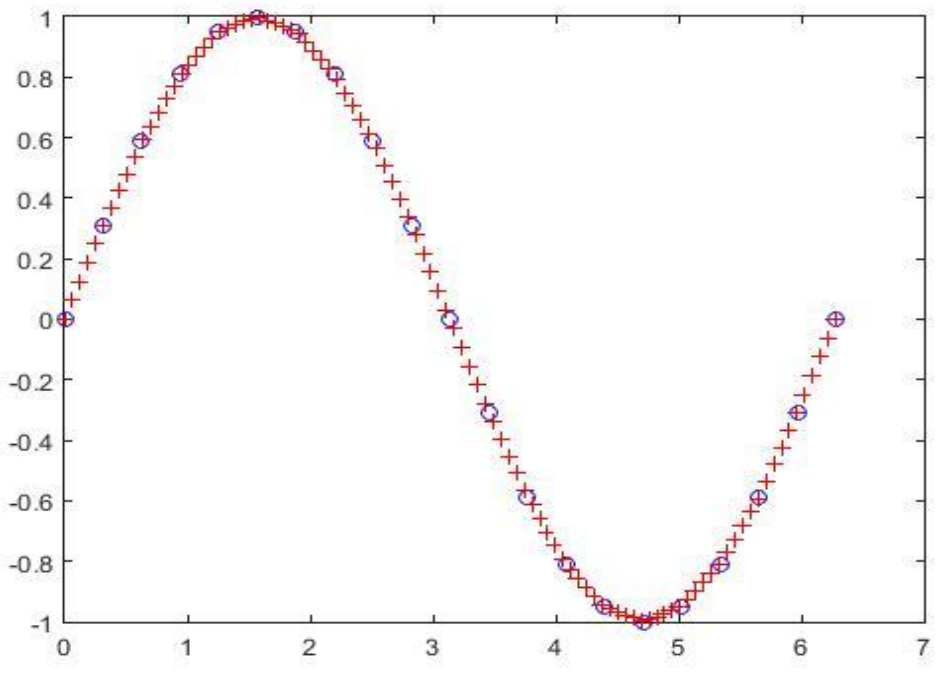

Figure 4-2 The effect drawing of ADB interpolation for function $y$=sin $x$

From the interpolation method of the univariate function and the method of finding the approximate value of the function of $n$ variables above, we obtain a general $n$-dimensional ADB interpolation method, that is: decompose an $n$-dimensional interpolation into $n$ one-dimensional interpolation, then use one-dimensional ADB interpolation to find the corresponding approximate values, respectively, and finally synthesize the $n$ approximate values by using sum-times-difference formula into one value as the approximation value of the function of *n* variables. The *n* pairs of one-dimensional interpolation formulas are required for *n*-dimensional ADB interpolation, they are as follows:

$$\begin{cases} y=\dfrac{y_{ij\ldots s}-y_{i-1,j,\ldots,s}}{x_{1_i}-x_{1_{i-1}}}x_1+\dfrac{x_{1_i}y_{i-1,j,\ldots,s}-x_{1_{i-1}}y_{ij\ldots s}}{x_{1_i}-x_{1_{i-1}}}, & x_1\in[\dfrac{x_{1_i}+x_{1_{i-1}}}{2},\ x_{1_i}] \\ y=\dfrac{y_{ij\ldots s}-y_{i+1,j,\ldots,s}}{x_{1_i}-x_{1_{i+1}}}x_1+\dfrac{x_{1_i}y_{i+1,j,\ldots,s}-x_{1_{i+1}}y_{ij\ldots s}}{x_{1_i}-x_{1_{i+1}}}, & x_1\in[x_{1_i},\ \dfrac{x_{1_i}+x_{1_{i+1}}}{2}] \end{cases}$$

$$\begin{cases} y=\dfrac{y_{ij\ldots s}-y_{i,j-1,\ldots,s}}{x_{2_i}-x_{2_{i-1}}}x_2+\dfrac{x_{2_i}y_{i,j-1,\ldots,s}-x_{2_{i-1}}y_{ij\ldots s}}{x_{2_i}-x_{2_{i-1}}}, & x_2\in[\dfrac{x_{2_i}+x_{2_{i-1}}}{2},\ x_{2_i}] \\ y=\dfrac{y_{ij\ldots s}-y_{i,j+1,\ldots,s}}{x_{2_i}-x_{2_{i+1}}}x_2+\dfrac{x_{2_i}y_{i,j+1,\ldots,s}-x_{2_{i+1}}y_{ij\ldots s}}{x_{2_i}-x_{2_{i+1}}}, & x_2\in[x_{2_i},\ \dfrac{x_{2_i}+x_{2_{i+1}}}{2}] \end{cases}$$

$$\cdots \qquad \cdots \qquad \cdots$$

$$\begin{cases} y=\dfrac{y_{ij\ldots s}-y_{i,j,\ldots,s-1}}{x_{n_i}-x_{n_{i-1}}}x_n+\dfrac{x_{n_i}y_{i,j,\ldots,s-1}-x_{n_{i-1}}y_{ij\ldots s}}{x_{n_i}-x_{n_{i-1}}}, & x_n\in[\dfrac{x_{n_i}+x_{n_{i-1}}}{2},\ x_{1_i}] \\ y=\dfrac{y_{ij\ldots s}-y_{i,j,\ldots,s+1}}{x_{n_i}-x_{n_{i+1}}}x_n+\dfrac{x_{n_i}y_{i,j,\ldots,s+1}-x_{n_{i+1}}y_{ij\ldots s}}{x_{n_i}-x_{n_{i+1}}}, & x_n\in[x_{n_i},\ \dfrac{x_{n_i}+x_{n_{i+1}}}{2}] \end{cases}$$

where $y_{ij\ldots s}$ is the value of function at base point $\left(x_{1_i}, x_{2_j}, \ldots, x_{n_s}\right)$, that is, $y_{ij\ldots s}= f(x_{1_i}, x_{2_j}, \ldots, x_{n_s})$. And the final synthesis formula (i.e., sum-times-difference formula) of approximate value of the function is

$$y=\sum_{i=1}^{n} y_{x_i}-(n-1)y_{ij\ldots s} \tag{4-12}$$

here $y_{x_i}$ is the approximate value of function that obtained by one-dimensional ADB interpolation for $x_i$ ($i$=1,2,…,$n$).

Actually, it is not difficult to see that the equation (4-12) can also be said to be the formula of multidimensional ADB interpolation, which is also a calculation model of high-dimensional interpolation.

**Example** 4-2 Using ADB interpolation to do interpolation for function $z=\frac{1}{4}x^2-\frac{1}{4}y^2$, the effect is shown in Figure 4-3.

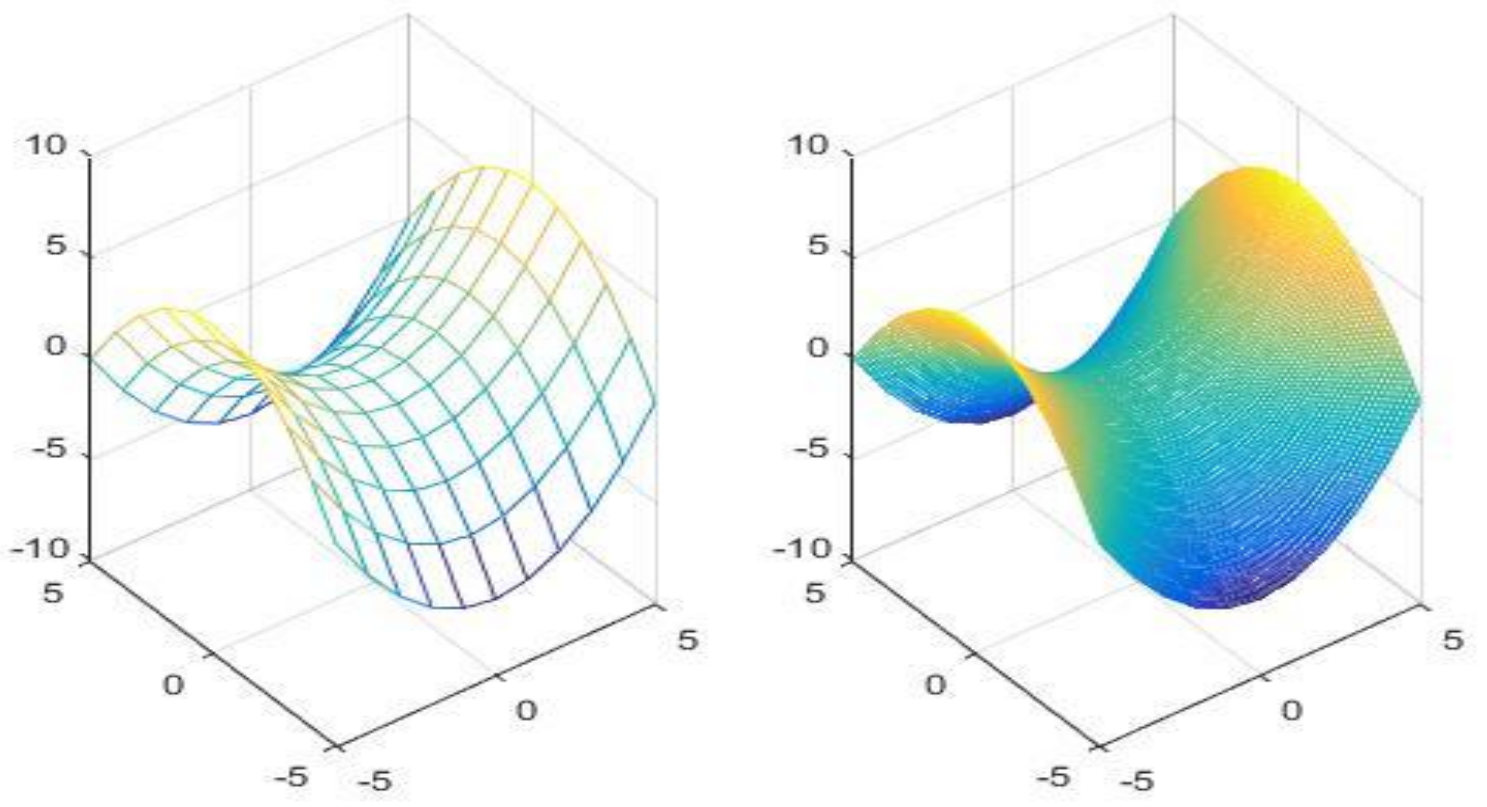

(a) (b)

Figure 4-3 The effect drawing of ADB interpolation for function $z = -x^2-y^2$

Where (a) is the functional graph before interpolating, and (b) is the functional graph after interpolating

**Example** 4-3 Using ADB interpolation to do interpolation for function of three variables, $u=ze^{-x^3-y^3-z^3}$, the effect (slice chart) is shown in Figure 4-4.

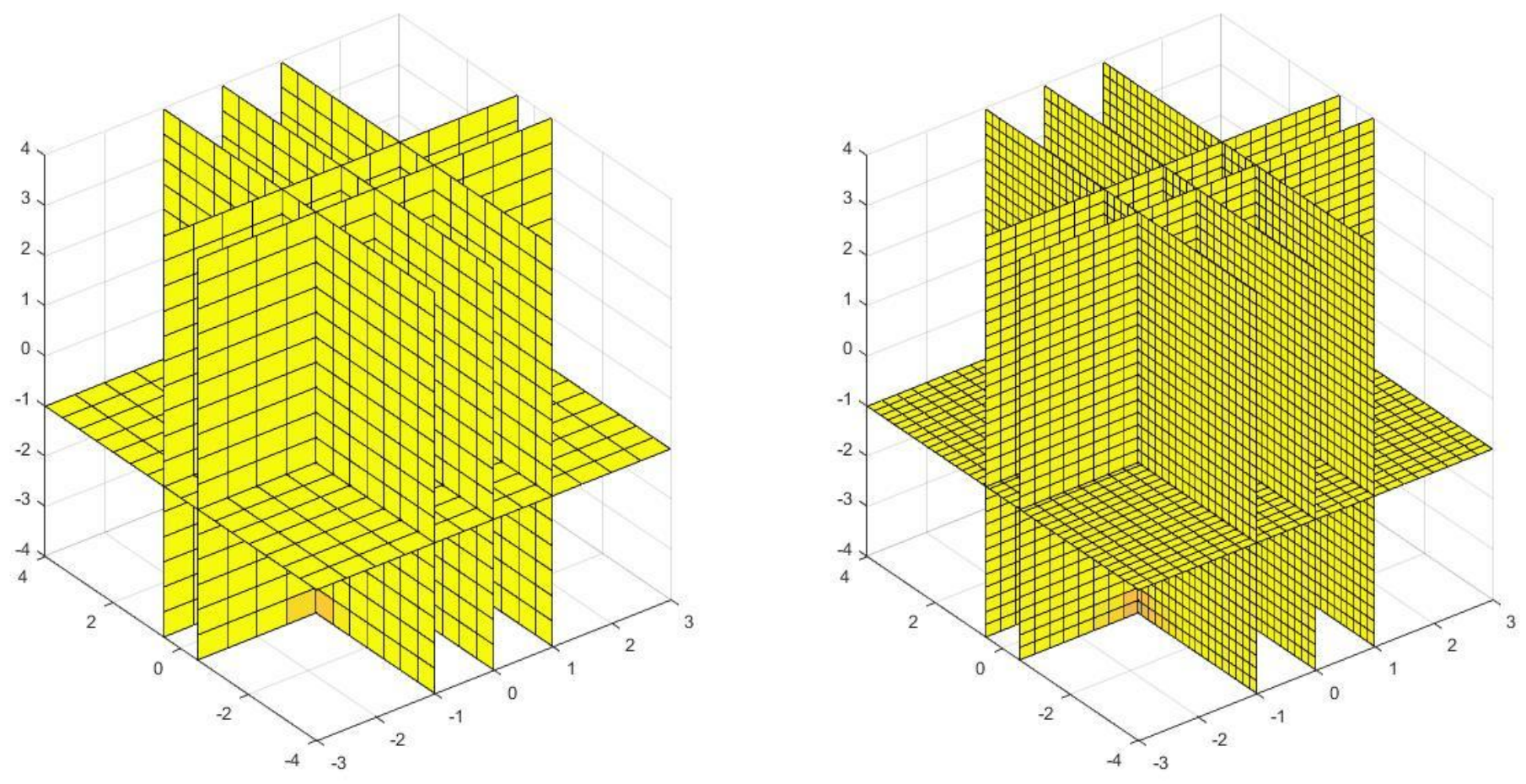

(a) (b)

Figure 4-4 The effect drawing (slice chart) of ADB interpolation for function of three variables, $u=ze^{-x^3-y^3-z^3}$

Where (a) is the functional graph before interpolating, and (b) is the functional graph after interpolating

(Note: due to the space limit, ADB interpolation with scattered data points will be introduced in another article.)

## 5 Instance-Based Learning Using ADB interpolation

In the above, we develop a new interpolation approach, ADB interpolation, from finding approximate value of a function. Since ADB interpolation is a local interpolation, it can be used for instance-based machine learning. In the following, we present two learning algorithms.

**(1) A leaning algorithm for regression problems** (in which training examples are regularly distributed)

---

- Put samples in sample set $X=\{(\boldsymbol{x}^i, f(\boldsymbol{x}^i))\}_{i=1}^{m}$ ($\boldsymbol{x}=(x_1, x_2 \dots , x_n)$) into a corresponding data structure $S$ in centralized or distributed manner (as training examples);
- For the currently being queried $\boldsymbol{x}'=(x_1', x_2', \dots , x_n')$:
  - According to its coordinate components $x_1', x_2', \dots , x_n'$ look up in $S$ sequentially or in parallel to determine a $\boldsymbol{x}^k$ ($k\in\{1,2,\dots,m\}$) to which the approximation-degree of $\boldsymbol{x}'$ is highest;
  - Take $\boldsymbol{x}^k$ as the center, and according to the position of $\boldsymbol{x}'$ relative to $\boldsymbol{x}^k$, choose $n$ corresponding nearest neighbors $\boldsymbol{x}^1, \boldsymbol{x}^2, \dots , \boldsymbol{x}^n$ (The data points here are renumbering.) from $S$, then construct the corresponding one-dimensional interpolation formulas

$$y_{x_j} = g(x_j| x_j^k, f(\boldsymbol{x}^k), \ x_j^l, f(\boldsymbol{x}^l)) \qquad (j=1,\dots, n; \ l\in\{1,2,\dots,n\})$$

and then compute $y_{x_1}, y_{x_2}, \dots, y_{x_n}$ sequentially or in parallel;

• Return

$$\hat{f}(\boldsymbol{x}')=\sum_{j=1}^{n} y_{x_j}-(n-1)\, f(\boldsymbol{x}^k)$$

---

**Example** 5-1 Take the following pairs of corresponding values of function $z=-x^2-y^2$, $((x_i, y_j), z_{ij})$ , as example data:

$x_i$: −20, −18, −16, … , −2, 0, 2, … , 16, 18, 20.

$y_j$: −20, −18, −16, … , −2, 0, 2, … , 16, 18, 20.

$z_{ij}$ : $-x_i^2-y_j^2$

and take the following data points, $(x, y)$, as query data:

$x$: −20, −20, 20, −19.5, −17.8, -18,−15.3, −12, −10.2, −10, −10, 0, 0, 10, 10, 5.6, 4.7, −3.4, −1.8, −2.3, −3.6, 1.2, −5.4, −15.6, −20, −20, −20, −18.3, 18.4, 17.5, 16.2, 14.5, 11.1, −5.4, −12.1, −8.5, −13.9, −7.5, −7.8, −9.8, −12.4, −13.5, −14.6, −17.5, −17.8.

$y$: −20, 20, −20, −19.5, −17.8,−5, −15.5, 2.5, −10.2, 10, −20, 0, −20, −20, −10, −15.3, −3.8, −13.4, −2.8, −1.9, −5.6, −10.2, −6.5, 5.6, 0, −10, 10, 10.4, −18.1, −16.3, −14.4, −12.3, −6.3, −15.8, −8.2, −15.6, 0.9, 1.6, 3.2, 4.6, 6.6, 2.8, −0.9, 18.6, 13.2.

we use the above learning algorithm, the corresponding $z$-values obtained are as follows:

| | | | | | | |
|---|---|---|---|---|---|---|
| −800.0000 | −800.0000 | −800.0000 | −762.0000 | −634.4000 | −350.0000 | −476.0000 |
| −151.0000 | −208.8000 | −200.0000 | −500.0000 | 0 | −400.0000 | −500.0000 |
| −200.0000 | −267.0000 | −37.8000 | −192.8000 | −12.4000 | −9.6000 | −45.6000 |
| −106.8000 | −73.0000 | −276.0000 | −400.0000 | −500.0000 | −500.0000 | −444.2000 |
| −667.0000 | −573.2000 | −470.8000 | −362.8000 | −164.4000 | −280.0000 | −214.2000 |
| −317.0000 | −195.2000 | −60.2000 | −72.4000 | −118.4000 | −198.8000 | −191.8000 |
| −215.8000 | −653.8000 | −492.4000 | | | | |

The effect drawing is shown in Figure 5-1.

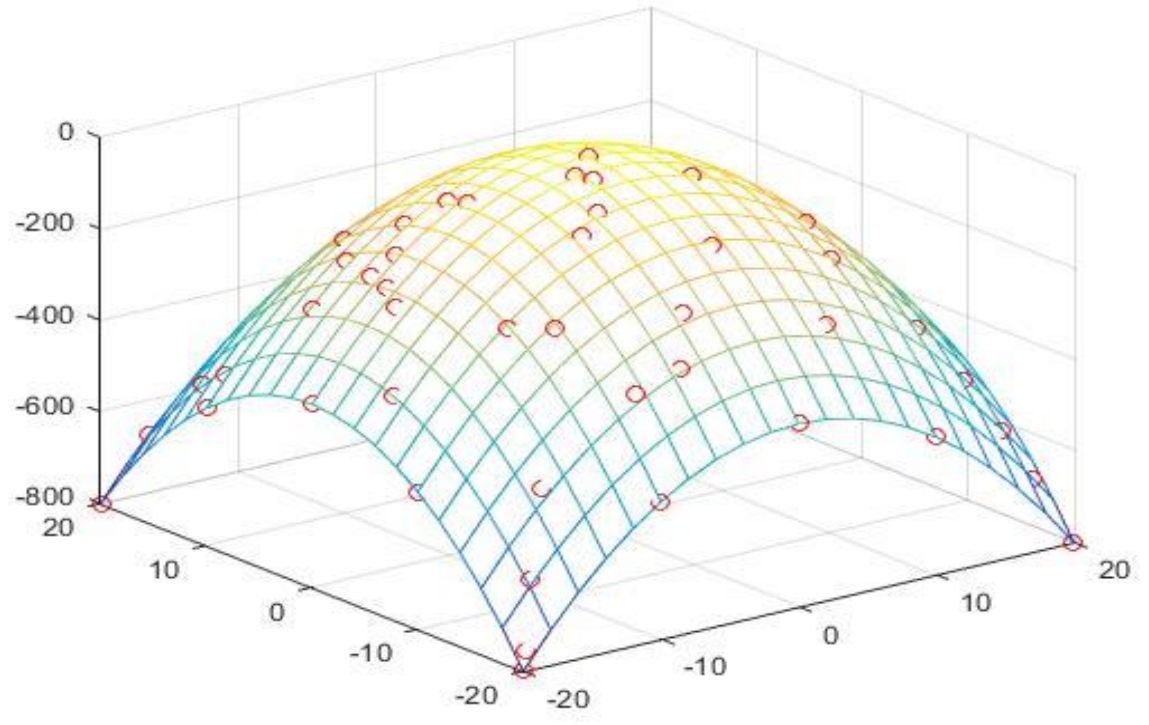

Figure 5-1 An effect drawing of the learning using ADB interpolation

Where the grid curve is the graph formed by example data from function $z=-x^2-y^2$, and the red circles indicate the points obtained by learning using ADB interpolation.

**Example** 5-2 Figure 5-2 below shows an effect drawing of learning using ADB interpolation. The example data are taken from the peaks function in MATLAB, and the query data is also designed according to this function. Limited by space, these data are omitted.

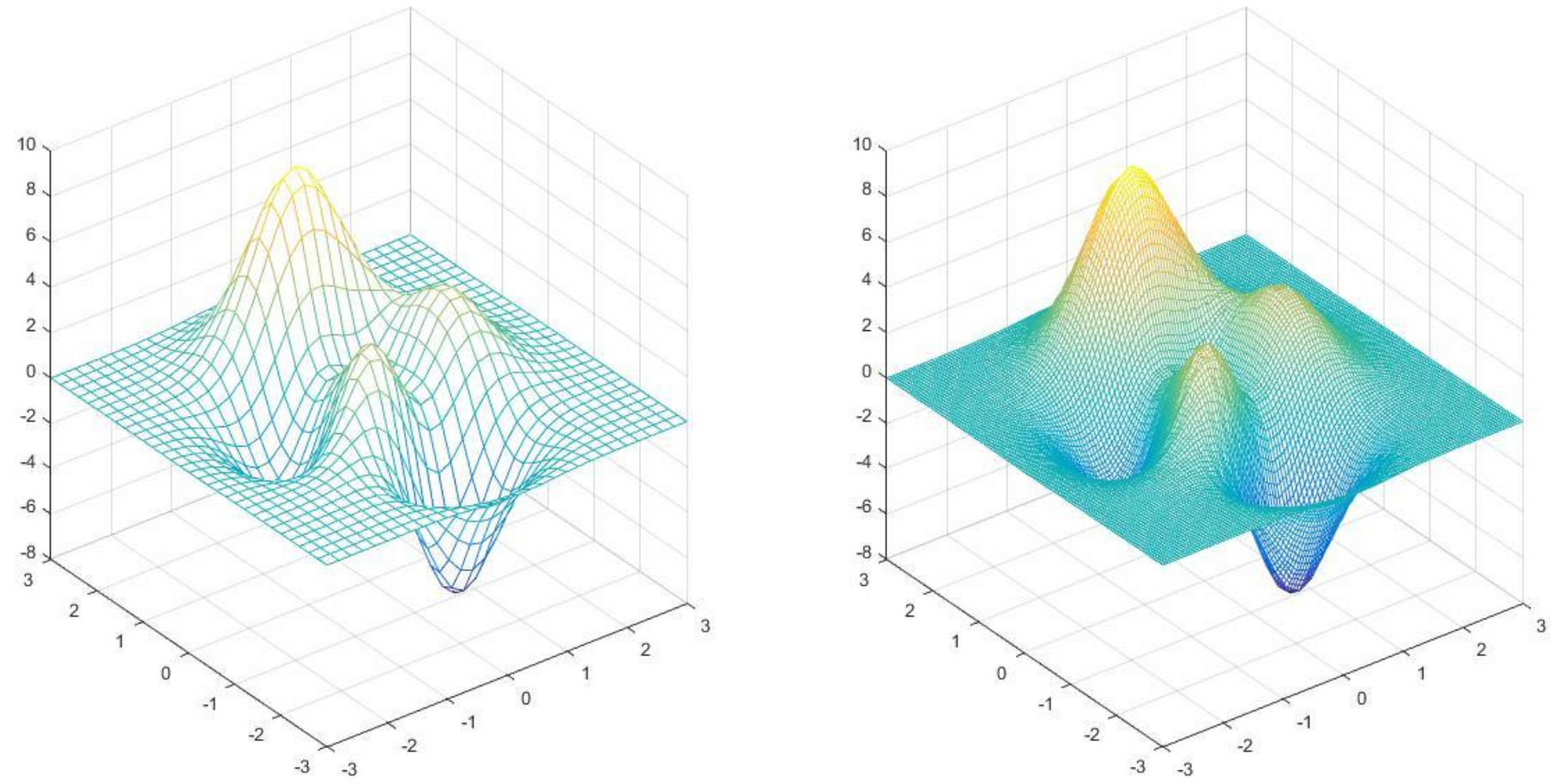

Figure 5-2  An illustration of the effect of learning using ADB interpolation

Where the left is the functional graph formed by example data, and the right is the functional graph obtained by learning using ADB interpolation.

As can be seen from the above, this instance-based learning using ADB interpolation has the following characteristics:

• The learning method takes data points ($\boldsymbol{x}^l$) of training examples (($\boldsymbol{x}^l$, $f(\boldsymbol{x}^l)$) as centers to set approximation regions and compute approximation-degrees.

• ADB interpolation is a local interpolation, the training examples involved in interpolation are related to the position of the currently queried data point $\boldsymbol{x}'$ relative to its nearest data point $\boldsymbol{x}^k$, the number of which is related to the dimension of the vector $\boldsymbol{x}$, $n$-dimensional $\boldsymbol{x}$ only involves 1+$n$ training examples (($\boldsymbol{x}^k$, $f(\boldsymbol{x}^k)$) and ($\boldsymbol{x}^1$, $f(\boldsymbol{x}^1)$), ($\boldsymbol{x}^2$, $f(\boldsymbol{x}^2)$), … , ($\boldsymbol{x}^n$, $f(\boldsymbol{x}^n)$) ). But since the point $\boldsymbol{x}'$ is only approximate to the point $\boldsymbol{x}^k$, the corresponding $y_{x_j}$ ($j$=1,…, $n$) are most affected by the example ($\boldsymbol{x}^k$, $f(\boldsymbol{x}^k)$), and the final synthesized value $\hat{f}(\boldsymbol{x}')$ is also most affected by ($\boldsymbol{x}^k$, $f(\boldsymbol{x}^k)$).

• If distributed storage and parallel processing (including parallel lookup and parallel computation) are used, the time complexity of corresponding algorithm is independent of the dimension of vector $\boldsymbol{x}$, and its efficiency is almost equal to that of one-dimensional interpolation at all time.

• The interpolation formulas (including sum-times-difference formula) are derived entirely by the mathematical method, so they have definite mathematical basis.

• The sum-times-difference formula is actually a linear combination of coordinate components of an interpolation point, and the denominators of the coefficients before each coordinate component are separately the difference between the coordinate component and the corresponding coordinate component of corresponding base point, that is, the distance between the two, so, these coefficients happen to also have a function of the weight values. Thus, viewed from the form, the sum-times-difference formula is a linear weighted regression formula. That is to say, the sum-times-difference formula here coincides with the traditional local weighted linear regression model. However, in local weighted linear regression, these coefficients are determined by searching, i.e., learning, while in our ADB interpolation, these coefficients are determined by looking up. The former is guided and constrained by error function (e.g. $E=\frac{1}{2}\sum_{\boldsymbol{x}\in X_1\subset X}(f(\boldsymbol{x})-\hat{f}(\boldsymbol{x}))^2$), and the latter by approximation-degree function (e.g. $A_{x_i}(x)$). In the sense of approximation-degree, the approximate value $\hat{f}(\boldsymbol{x}')$ of the function obtained from the sum-times-difference formula is always the most accurate.

• The accuracy of the returned approximate value $\hat{f}(\boldsymbol{x}')$ of a function is positively related to the approximation-degree of $\boldsymbol{x}'$ to $\boldsymbol{x}^k$.

• Comparing the learning using ADB interpolation with the deep learning, the deep learning is to approximate objective function with the strategy of deepening vertically[9], yet the learning using ADB interpolation can approximate objective function with the strategy of increasing density horizontally. So, the learning using ADB interpolation and the deep learning can complement each other, and can even have an effect of "different approaches but equal results" in the case of sufficient samples.

**(2) A learning algorithm for classification problems**

---

- Put samples in sample set $X=\{(\boldsymbol{x}^i, f(\boldsymbol{x}^i))\}_{i=1}^m$ ($\boldsymbol{x}=(x_1, x_2, \ldots, x_n)$, $f(\boldsymbol{x})$ is a class label ) into a corresponding data structure $S$ in centralized or distributed manner; (as training examples)
- For the currently being queried $\boldsymbol{x}'=(x_1', x_2', \ldots, x_n')$:
  - According to its coordinate components $x_1', x_2', \ldots, x_n'$ look up in $S$ sequentially or in parallel to determine a $\boldsymbol{x}^k$ ($k\in\{1,2,\ldots,m\}$) to which the approximation-degree of $\boldsymbol{x}'$ is highest;
  - If such a $\boldsymbol{x}^k$ is found, return

$$\hat{f}(\boldsymbol{x}')=f(\boldsymbol{x}^k)$$

  Else exit.

---

**Example** 5-3 Suppose that the data points of training examples of a classification problem are shown as the black circles in Figure 5-3, and the small boxes surrounding them are their respective strict approximation regions; and the white circles represent data points to be classified. Using the learning algorithm that using ADB interpolation to classify these data points, the classifying results are shown in the figure. As you can see, there are two queried data points are respectively classified to two classes to which the corresponding training example data points belong, because they are located respectively in the strict approximation regions of the corresponding data points, while the two queried data points outside the small boxes are not classified.

It can be seen that for classification problems, the learning using ADB interpolation has the following advantages:

• Although similar to the traditional 1-NN algorithm, the approximation regions in the algorithm are aimed at the data points of training examples, and which are strict approximation regions of square.

• Since the approximation and approximation regions involved in the algorithm are strict approximation and strict approximation regions, for classification problems, learning using ADB interpolation avoids, from the source, the problem in traditional instance-based learning (e.g. $k$-NN algorithm) that some datum objects with similar partial attributes are classified into a class.

• The classifying result of learning using ADB interpolation is unique, and there is no the phenomenon like $k$-NN algorithm, that classifying result may change with the change of $k$'s value.

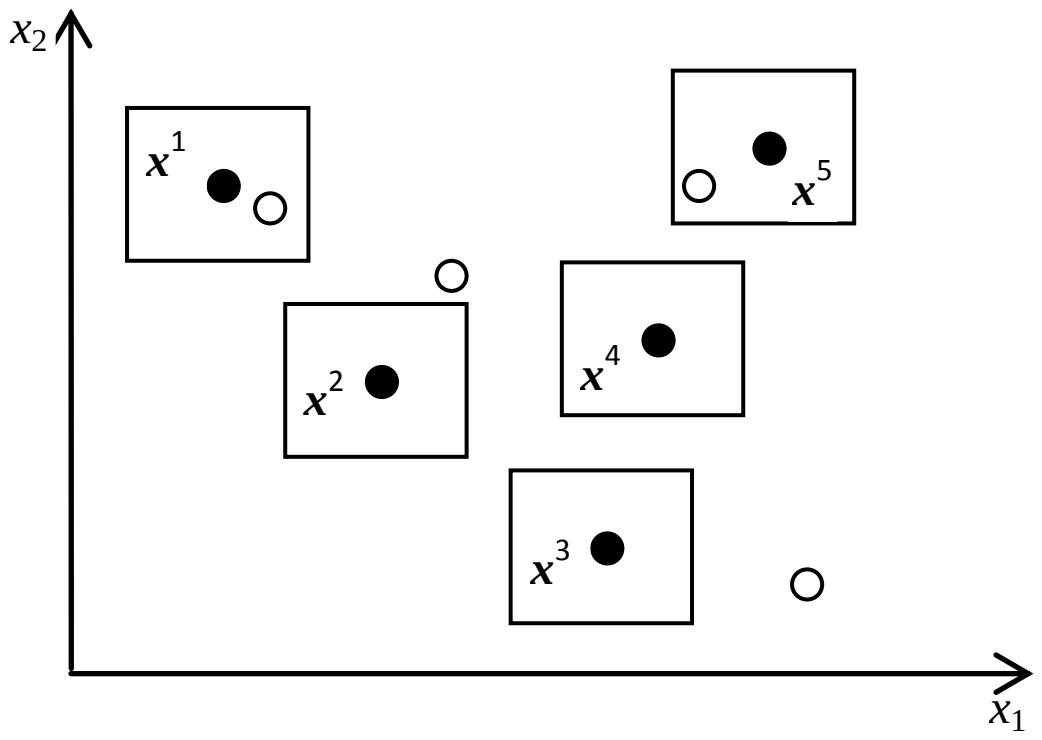

Figure 5-3 An illustration of applying the learning using ADB interpolation to classification

## 6 Summary

In this paper, started from finding approximate value of a function, we introduced the measure of approximation-degree between two numerical values, proposed the concepts of "strict approximation" and "strict approximation region", then, derived the corresponding one-dimensional interpolation methods and formulas, and then presented a calculation model called "sum-times-difference formula" for high-dimensional interpolation, thus we developed a new interpolation approach — approximation-degree-based interpolation, i.e., ADB interpolation. ADB interpolation was applied to the interpolation of actual functions with satisfactory results. Viewed from principles and examples, the approach is of novel idea, and it has the advantages of simple calculations (they are all arithmetic), stable accuracy (benefitting from local interpolation and that the approximation-degrees are always not less than 0.5); especially, the approach facilitates parallel processing, very suiting for high-dimensional interpolation, and easy to be extended to the interpolation of vector valued functions.

Applying ADB interpolation to instance-based learning, we obtained a new instance-based learning method — learning using ADB interpolation, and we also gave several examples of the learning. Viewed from principles and examples, the learning method is of unique technique (e.g., taking data points of training examples as centers to set approximation regions that are also strict approximation regions and compute approximation-degrees). Besides the advantages of ADB interpolation, the learning method has also the advantages of definite mathematical basis, implicit distance weights, avoiding misclassification (guaranteed by strict approximation), high efficiency (benefitting from distributed storage and parallel processing), and wide range of applications (which can be applied to the regression or classification problems, and can be used for large sample learning or small sample even single sample learning), as well as being interpretable, etc. In principle, this method is a kind of learning by analogy, which and the deep learning that belongs to inductive learning can complement each other, and for some problems, the two can even have an effect of “different approaches but equal results” in big data and cloud computing environment. Thus, the learning using ADB interpolation can also be regarded as a kind of “wide learning” that is dual to deep learning.